\documentclass[a4paper]{cas-dc}  

\usepackage[authoryear]{natbib}
\usepackage{amsmath,amsfonts}
\usepackage{amsthm}
\usepackage{algorithmic}
\usepackage{threeparttable}
\usepackage[switch]{lineno}
\usepackage[ruled,linesnumbered]{algorithm2e}

\def\tsc#1{\csdef{#1}{\textsc{\lowercase{#1}}\xspace}}
\tsc{WGM}
\tsc{QE}
\hypersetup{breaklinks=false,linkcolor=red,anchorcolor=black,colorlinks}

\begin{document}
\let\WriteBookmarks\relax
\def\floatpagepagefraction{1}
\def\textpagefraction{.001}
\let\printorcid\relax 
\shorttitle{Predicting Gradient is Better: Exploring Self-Supervised Learning for SAR ATR with a Joint-Embedding Predictive Architecture}   

\shortauthors{Weijie Li et al.}

\title[mode = title]{\textcolor{black}{Predicting Gradient is Better: Exploring Self-Supervised Learning for SAR ATR with a Joint-Embedding Predictive Architecture}}  

\tnotetext[1]{This document was \textcolor{black}{partially} supported by the National Key Research and Development Program of China (No. 2021YFB3100800), the National Natural Science Foundation of China (No. 61871384, \textcolor{black}{62376283, 62022091, 62201588}, and 61921001), the Science and Technology Innovation Program of Hunan Province (No. 2022RC1092), and \textcolor{black}{the Key Stone Grant JS2023-03 of the National University of Defense Technology.}}

\author[1]{Weijie Li}[style=chinese]
\ead{lwj2150508321@sina.com} 

\author[1]{Wei Yang}[style=chinese]
\ead{yw850716@sina.com} 

\author[1]{Tianpeng Liu}[style=chinese]
\ead{everliutianpeng@sina.cn} 

\author[2]{Yuenan Hou}[style=chinese]
\ead{houyuenan@pjlab.org.cn} 

\author[3]{Yuxuan Li}[style=chinese]
\ead{yuxuan.li.17@ucl.ac.uk} 

\author[1]{Zhen Liu}[style=chinese]
\ead{zhen_liu@nudt.edu.cn} 

\author[1]{Yongxiang Liu}[style=chinese]
\cormark[1]
\ead{lyx_bible@sina.com} 

\author[1]{Li Liu}[style=chinese]
\ead{liuli_nudt@nudt.edu.cn} 

\address[1]{the College of Electronic Science and Technology, National University of Defense Technology, Changsha, 410073, China}
\address[2]{the Shanghai AI Laboratory, Shanghai, 200000, China}
\address[3]{the Visual Computing and Intelligent Perception Lab, Nankai University, Tianjin, 300071, China}

\cortext[1]{Corresponding author: Li Liu, Wei Yang, Tianpeng Liu and Yongxiang Liu} 

\begin{abstract}
The growing Synthetic Aperture Radar (SAR) \textcolor{black}{data can build a foundation model using self-supervised learning (SSL) methods, which can achieve various SAR automatic target recognition (ATR) tasks with pretraining in large-scale unlabeled data and fine-tuning in small-labeled samples. SSL aims to construct supervision signals directly from the data, minimizing the need for expensive expert annotation and maximizing the use of the expanding data pool for a foundational model. This study investigates an effective SSL method for SAR ATR, which can pave the way for a foundation model in SAR ATR.} The primary obstacles faced in SSL for SAR ATR are small targets in remote sensing and speckle noise in SAR images, \textcolor{black}{corresponding to the SSL approach and signals}. To overcome these challenges, we present a novel \textcolor{black}{joint-embedding predictive architecture for SAR ATR (SAR-JEPA) that }leverages local masked patches to predict the \textcolor{black}{multi-scale SAR gradient representations} of an unseen context. The key aspect of \textcolor{black}{SAR-JEPA} is integrating SAR domain features to \textcolor{black}{ensure high-quality self-supervised signals as target features}. \textcolor{black}{In addition}, we employ local masks and multi-scale features to \textcolor{black}{accommodate various small targets} in remote sensing. By \textcolor{black}{fine-tuning and evaluating} our framework on three target recognition datasets (vehicle, ship, and aircraft) \textcolor{black}{with four other datasets as pretraining}, we demonstrate its outperformance over other SSL methods and its effectiveness \textcolor{black}{as the} SAR data \textcolor{black}{increases}. This study \textcolor{black}{demonstrates} the potential of SSL for \textcolor{black}{the recognition of SAR targets} across diverse targets, scenes, and sensors. \textcolor{black}{Our codes and weights are available in \url{https://github.com/waterdisappear/SAR-JEPA}.}
\end{abstract}



\begin{keywords}
Synthetic Aperture Radar (SAR)\sep 
Automatic Target Recognition (ATR)\sep 
Self-Supervised Learning (SSL)\sep 
Deep Learning\sep 
Masked Image Modeling (MIM)
\end{keywords}

\maketitle

\section{Introduction}
\label{Introduction}
Synthetic aperture radar (SAR) \textcolor{black}{can} capture images under \textcolor{black}{various} weather and lighting conditions, \textcolor{black}{making} it indispensable for acquiring Earth observation information~\citep{sun2021spaceborne,ref1,tsokas2022sar}. As a crucial component in \textcolor{black}{the} interpretation \textcolor{black}{of the SAR images}, SAR automatic target recognition (SAR ATR) aims to localize and classify the desired targets within SAR images. Over the past few decades, extensive \textcolor{black}{studies} have been conducted in this field with various civilian and military applications, including transportation management~\citep{gagliardi2023satellite}, autonomous driving~\citep{rizzi2021navigation}, and military surveillance~\citep{10283916, 9915465}. \textcolor{black}{Deep learning has revitalized SAR ATR in the last decade with remarkable advancements~\citep{kechagias2021automatic, li2023comprehensive,datcu2023explainable}}. Despite significant progress \textcolor{black}{made}, the high costs of data collection and annotation limit dataset diversity and method applications. SAR ATR studies are \textcolor{black}{classified} into subfields, such as vehicles~\citep{10283916, peng2024towards}, ships~\citep{hou2020fusar}, and aircraft recognition~\citep{zhao2022attentional}, \textcolor{black}{using many specialized approaches}. Hence, an intriguing and worthwhile question to explore is~\emph{whether a generic representation and method exist for various SAR target recognition tasks}.

Emerging self-supervised learning (SSL) techniques present a potential \textcolor{black}{answer} to this question. SSL aims to extract meaningful signals directly from the data, thereby reducing the need for costly expert annotations and efficiently using \textcolor{black}{an increasing amount} of data. Foundation models trained on extensive datasets using SSL have demonstrated their effectiveness in various computer vision~\citep{goldblum2023battle,he2022masked}, remote sensing~\citep{9844015}, and clinical medical tasks~\citep{zhou2023foundation}. \textcolor{black}{Therefore}, we believe that foundation models, trained \textcolor{black}{using} SSL on a large dataset, can \textcolor{black}{produce} a generalized representation for SAR target recognition. \textcolor{black}{Consequently}, this \textcolor{black}{study} investigates the key challenges in SSL for SAR ATR.

\begin{figure*}[!tb]
\centering
\includegraphics[]{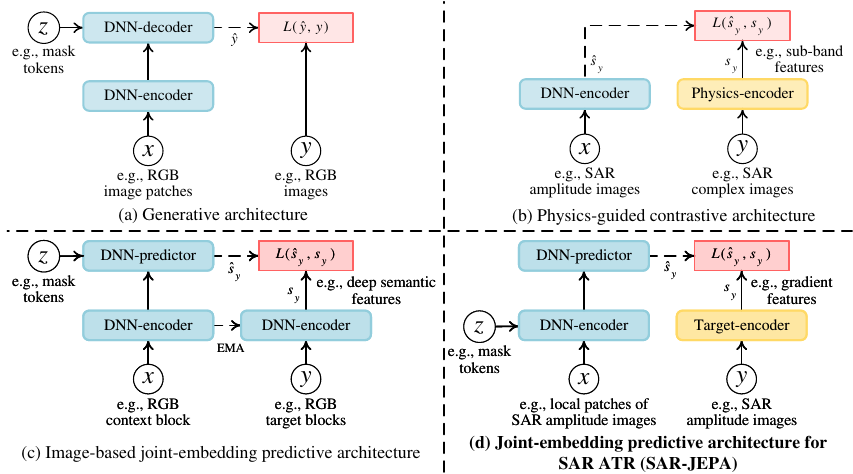}
\caption{\textcolor{black}{Comparison of related architectures and the proposed architecture.}
(a) Generative architecture~\citep{he2022masked} focuses on reconstructing the pixels of unseen patches with a high mask proportion. This approach creates a challenging and meaningful pretext task that allows the model to learn contextual relationships within images. 
(b) Physics-guided contrastive architecture (PGCA)~\citep{datcu2023explainable,huang2022physically} leverages the unique representation in the SAR domain as guided signals to constrain the representation of the deep neural network (DNN). By incorporating physical principles into the learning process, PGCA \textcolor{black}{improves} the ability \textcolor{black}{of the model} to capture accurate and important SAR features.
(c) \textcolor{black}{Image-based joint-embedding predictive architecture (I-JEPA)~\citep{assran2023self} uses deep features as target signals and \textcolor{black}{learns} more semantic information. However, we found that a learnable target encoder is susceptible to feature collapses due to SAR image noise.} 
(d) \textcolor{black}{Our joint-embedding predictive architecture for SAR ATR (SAR-JEPA)} combines the SAR domain \textcolor{black}{embedding} and the meaningful masked image modeling (MIM) pretext task to learn contextual relationships in the SAR gradient feature space. This approach utilizes prior knowledge about SAR target recognition \textcolor{black}{on} target scale and features to improve representation.}
\label{fig2_motivation}
\end{figure*}

\textbf{Lack of a large-scale pretraining set.} 
Just as ImageNet~\citep{fei2022searching} has achieved an impressive contribution to computer vision by \textcolor{black}{increasing} the number of categories for object recognition from 20 to thousands, a large-scale pretraining set for SAR target recognition can introduce new research ideas and tasks. Unfortunately, \textcolor{black}{owing} to the acquisition cost and annotation difficulties, many SAR target datasets have at most ten annotation categories and a few thousand images. \textcolor{black}{Although} some studies~\citep{huang2022physically,zhai2022weakly,10269665} \textcolor{black}{on} SAR target recognition used \textcolor{black}{some} datasets as pretraining and evaluations, the potential of SSL in large-scale SAR \textcolor{black}{datasets has not yet been fully explored}. Therefore, it is necessary to integrate increasing open-source datasets to construct a large-scale pretraining set \textcolor{black}{that contains various} targets, scenes, and sensors\footnote{Various imaging conditions regarding target, scene, and sensor can lead to complex target and background signature variations in SAR images~\citep{Ross1999SAR,10283916}.}.

\textbf{Special properties of SAR images.} 
SAR achieves coherent imaging \textcolor{black}{using} a moving platform \textcolor{black}{in remote sensing}. Its images reflect the electromagnetic scattering of objects, but they are often affected by multiplicative speckle noise, phase errors, and other interferences. Consequently, SAR image interpretation is extremely difficult and poses difficulties for SSL methods \textcolor{black}{because of their unique imaging properties}. Our study focuses on addressing two key issues: the scale problem and target features. First, remote sensing images often contain large scenes and small targets, making it difficult to learn effective contextual information for various targets. Current pretraining methods for SAR target recognition focus insignificantly on this issue because they mainly use a few datasets that lack significant scale problems in the dataset. Second, SAR images are affected by speckle noise, which can affect the quality of learned representations. This issue was addressed through data augmentation~\citep{zhai2022weakly} and filtering methods~\citep{10269665} in contrastive SSL, as well as the histogram of oriented gradient (HOG) features~\citep{wang2023feature} in generative SSL. However, data augmentation in contrastive learning cannot exactly simulate the effects of speckle noise, and the gradient computation in HOG features has \textcolor{black}{false results} due to multiplicative noise. Based on the above discussions, these two issues still require adequate solutions. We aim to leverage the knowledge of the SAR domain to address these unique challenges of SSL \textcolor{black}{in pretext tasks and target signals}.

\textcolor{black}{To} deploy SSL for SAR ATR, we first constructed a pretraining dataset \textcolor{black}{comprising} approximately 100,000 SAR magnitude images sourced from \textcolor{black}{four open-source} datasets~\citep{xia2022crtranssar,chen2022large,wang2019sar,malmgren2017improving,kusk2016synthetic,lewis2019sar}. This \textcolor{black}{pretraining} dataset encompasses a wide range of target categories, scenes, and sensors, allowing us to explore effective SSL methods. To address the \textcolor{black}{aforementioned} issues, we proposed a novel joint-embedding predictive architecture for SAR ATR (\textcolor{black}{SAR-JEPA}) in Figure~\ref{fig2_motivation}. \textcolor{black}{SAR-JEPA leverages gradient embedding and the masked image modeling (MIM) task to learn contextual relationships within the SAR feature space}. We employed the well-established gradient-by-ratio (GR)~\citep{dellinger2014sar,song2016sar} method as the target feature, \textcolor{black}{thereby} effectively mitigating the \textcolor{black}{effect} of speckle noise in SAR images and facilitating accurate target shape extraction. Furthermore, we introduced local masks and multi-scale features to enhance the adaptability of the architecture to \textcolor{black}{various small target scales} in remote sensing. To evaluate the effectiveness \textcolor{black}{of the proposed architecture}, we conducted experiments on three datasets (vehicle, ship, and aircraft). 

\textcolor{black}{Numerous experimental results} demonstrate that \textcolor{black}{SAR-JEPA} achieves improved performance than other SSL methods. Importantly, it can increase with the volume of pretraining data, thereby showing the potential of SSL to build a foundation model to unify the SAR ATR across various targets, scenes, and sensors. The main contributions of this \textcolor{black}{study} are as follows.

\begin{enumerate}
\item[$\bullet$] 
\textcolor{black}{We construct comprehensive SAR pretraining and fine-tuning settings to study SSL for SAR ATR self-supervised learning. A novel joint-embedding predictive architecture for SAR ATR (SAR-JEPA) is proposed to learn contextual relationships within the SAR gradient feature space.}
\item[$\bullet$] 
\textcolor{black}{The core of SAR-JEPA depends on the SSL method and target feature construction. SAR-JEPA learns contextual information around small targets in remote sensing images more efficiently by local masking. SAR-JEPA uses multi-scale gradient features as guidance signals to solve SAR speckle noise interference.}
\item[$\bullet$] 
\textcolor{black}{SAR-JEPA’s} performance with an increasing data volume demonstrates the potential of SSL to learn generalized feature representations. \textcolor{black}{We hope it will stimulate the enthusiasm for research on fundamental models} of SAR target recognition.
\end{enumerate} 

The \textcolor{black}{rest} of this paper is organized as follows. Section~\ref{Related Work} introduces related studies \textcolor{black}{on} SSL. Section~\ref{Approach} introduces the proposed SAR-JEPA. Section~\ref{Experiments} \textcolor{black}{presents} extensive experiments and analyses \textcolor{black}{of the effectiveness and scaling capabilities of our architecture}. Section~\ref{Conclusion} concludes the paper and discusses future work.

\section{Related Work}
\label{Related Work}
\subsection{Self-supervised learning \textcolor{black}{in computer vision}}
SSL~\citep{liu2021self, balestriero2023cookbook} aims to learn the intrinsic relationship in samples through a pretext task. This method can leverage a \textcolor{black}{large} amount of \textcolor{black}{unlabeled} data and perform well in various downstream tasks. SSL methods can be broadly \textcolor{black}{classified} into contrastive and generative approaches~\citep{liu2021self}. 
In this study, we focus on generative masked image modeling because it can model both local and long-range information~\citep{xie2023revealing}, and the data augmentation pretext task in the contrastive approach is not fully compatible with the \textcolor{black}{properties of the SAR image}. 
\textcolor{black}{Bidirectional encoder representation from image transformers} (BEiT)~\citep{bao2021beit} proposed the concept of masked image modeling and \textcolor{black}{predicted} masked patches in a frozen pretraining tokenized space. 
Compared to BEiT, the mask autoencoder (MAE)~\citep{he2022masked} achieved an efficient and meaningful task through the asymmetric autoencoder structure and the high masking rate. 
Furthermore, simple framework for MIM (SimMIM)~\citep{xie2022simmim} \textcolor{black}{has shown} that masked image modeling can be achieved with a simple structure, such as a linear projection layer and a random mask. Over time, several improvement studies have emerged, and our focus has been on research on masking methods and target features.

\textbf{Masking strategy} can affect the \textcolor{black}{quality of the representation}, which \textcolor{black}{focuses} on ensuring the difficulty of the pretext task and learning contextual information \textcolor{black}{around the} target regions. 
MAE~\citep{he2022masked} and SimMIM~\citep{xie2022simmim} compared different sampling strategies, such as random sampling, block sampling, and grid sampling, and found that random sampling with a large mask ratio worked the best. 
I-JEPA~\citep{assran2023self} employed block sampling and used a single context block to predict several nonoverlapping target blocks to exploit the rich semantics in the context block. 
Hard patch mining~\citep{wang2023hard} applied a teacher network to predict the reconstruction loss of different patches to better mine the target foreground region.
Pixel reconstruction in MIM (PixMIM)~\citep{liu2023pixmim} used a simple random resized crop to preserve more of the target foreground region in the random mask for \textcolor{black}{learning semantic concepts}.
Inspired by the fact that the target reconstruction in MAE uses mainly the surrounding area, Local masked reconstruction (LoMaR)~\citep{chen2022efficient} was proposed to perform masked reconstruction within a local image block instead of the entire image to improve computational efficiency. 
For remote sensing images, the masking strategy must consider their large scene and small target properties. We applied the LoMaR method to learn the local contextual information.

\textbf{Target features.} 
Masked image modeling learns contextual information by reconstructing and predicting various feature representations, such as pixel values~\citep{he2022masked, xie2022simmim}, discrete tokens~\citep{bao2021beit}, HOG features~\citep{wei2022masked, wang2023masked}, deep features~\citep{wei2022masked,assran2023self}, and frequency features~\citep{xie2022masked,liu2023pixmim}.
MaskFeat~\citep{wei2022masked} proved that the target features can be either the classical HOG features or the deep features obtained using other self-supervised methods. 
MIM with local multi-scale reconstruction (LocalMIM)~\citep{wang2023masked} performed multi-scale reconstruction of HOG features at different encoder layers to facilitate the representation learning speed and semantic understanding of scales. 
I-JEPA~\citep{assran2023self} used a learnable target encoder to obtain deep semantic features. 
PixMIM~\citep{liu2023pixmim} improved the learning of low-frequency components, such as target shape, by reconstructing the image after low-pass filtering.
Masked frequency modeling~\citep{xie2022masked} avoided spatially redundant masking by reconstructing various frequency components. 
A priori knowledge can be incorporated as a target feature to guide the representation learning. \textcolor{black}{In particular, our SAR-JEPA is inspired by the architecture of I-JEPA, which uses a learnable target encoder to learn deep semantic information. However, we found that this approach leads to feature collapse with SAR image noise.} Therefore, we used the GR method in SAR to extract the \textcolor{black}{features of the target shape and suppress} speckle noise. 

\subsection{Self-supervised learning in remote sensing}
In recent years, there has been \textcolor{black}{increasing} interest in SSL for remote sensing, which provides a paradigm for solving the contradiction between the increase in samples and the lack of high-quality annotations~\citep{wang2022self}. SSL has many applications in multi-source data, such as multispectral~\citep{reed2023scale,9844015}, hyperspectral~\citep{ibanez2022masked}, and SAR~\citep{wang2023feature,huang2022physically,zhai2022weakly}, \textcolor{black}{with various contrastive and generative paradigms}.

\textcolor{black}{Owing} to the lack of a large-scale dataset, SSL in SAR ATR is relatively scarce~\citep{zhai2022weakly, 10269665, 10283916,ref11}. These studies rely on contrastive learning or information on the target rotation angle. However, the data augmentation methods applied to natural images may not be fully suitable for SAR images~\citep{wang2022self}. For example, Gaussian blurring and noise methods cannot simulate multiplicative speckle noise effects, and the rotation is incompatible with SAR images because of the anisotropic electromagnetic scattering. In addition, only a few datasets provide information on the target rotation angle. Compared to applying multiple data augmentation methods to simulate imaging condition variations and noise interference for SAR images~\citep{zhai2022weakly, 10269665, 10283916}, we prefer another contrastive architecture~\citep{datcu2023explainable} compatible with the properties of the SAR image. Its contrastive architecture used \textcolor{black}{SAR domain features} to guide representation learning. We combine this idea with MIM to learn the contextual semantic information within the feature space. The target features with the SAR image properties can ensure a high-quality representation. We discuss SSL learning in remote sensing and SAR from three \textcolor{black}{viewpoints}.

\textbf{Pretrain dataset.} 
There is no unified benchmark setting for SSL with SAR target recognition owing to the lack of a large dataset, such as ImageNet. For scene classification~\citep{wang2023feature, huang2022physically}, the pretraining sets used BigEarthNet-MM~\citep{sumbul2021bigearthnet} or the sea ice dataset~\citep{huang2022physically}. 
For target recognition~\citep{zhai2022weakly,10269665}, the pretraining sets applied the moving and stationary target acquisition and recognition (MSTAR) vehicle~\citep{MSTAR} or OpenSARship~\citep{huang2017opensarship} datasets. Previous studies on SAR target recognition~\citep{zhai2022weakly,10269665} used pretraining sets that contained several target categories and scenarios. Therefore, the potential of SSL for foundation models in SAR target recognition \textcolor{black}{has not yet been} investigated. Constructing a larger dataset containing various targets, scenes, and sensors is necessary.

\textbf{Scale of target and scene.} 
A major difference between remote sensing and computer vision is in scene range and target size. Remote sensing images usually contain large scenes and small targets~\citep{li2023large,cheng2023towards}. 
The original vision (TOV)~\citep{10110958} illustrated that small scenes with different semantics within a larger scene affect contrastive learning. 
SAR-optical data fusion~\citep{9614157} used the shift alignment of two local regions between SAR and optical images in contrast learning.
Remote sensing foundation model framework (RingMo)~\citep{9844015} proposed a patch incomplete mask sampling strategy for \textcolor{black}{small dense} targets in remote sensing images, that is, secondary pixel sampling of the masked patches to expose small targets. 
However, we found that pixel sampling does not work for SAR images because the single pixels in the SAR images contain multiplicative noise. Therefore, we prefer local patches for masking instead of the entire image or the pixel level.

\textbf{SAR speckle noise.} 
Speckle noise affects the quality of the SSL features. \textcolor{black}{To address this problem, researchers have applied} many methods, such as data augmentation~\citep{zhai2022weakly}, filtering~\citep{10269665}, and feature extraction.
\textcolor{black}{Batch instance discrimination and feature clustering (BIDFC)}~\citep{zhai2022weakly} added Gaussian noise and blur to data augmentation methods. However, Gaussian noise and blurring do not perfectly match the multiplicative speckle noise.
\textcolor{black}{The despeckling preprocessing} was used in contrastive learning to address speckle noise~\citep{10269665}. 
\textcolor{black}{We focus on extracting high-quality features, such as guide signals, which accomplish noise reduction and feature extraction.}
Similar to our study, \textcolor{black}{feature guided masked autoencoder (FG-MAE)}~\citep{wang2023feature} discussed different features such as CannyEdge~\citep{canny1986computational}, HOG, and SAR scale-invariant feature transform (SAR-SIFT)~\citep{dellinger2014sar} features for SAR scene classification and segmentation. They found that HOG is more appropriate than SAR-SIFT for SAR SSL. This study aims to extract semantic information at the target level rather than at the scene level through local patches, \textcolor{black}{and we found that GR is more suitable for target shape information extraction than the differential gradient of HOG under multiplicative speckle noise.}

\section{Approach}
\label{Approach}
\begin{figure*}[!tb]
\centering
\includegraphics[width=17.5cm]{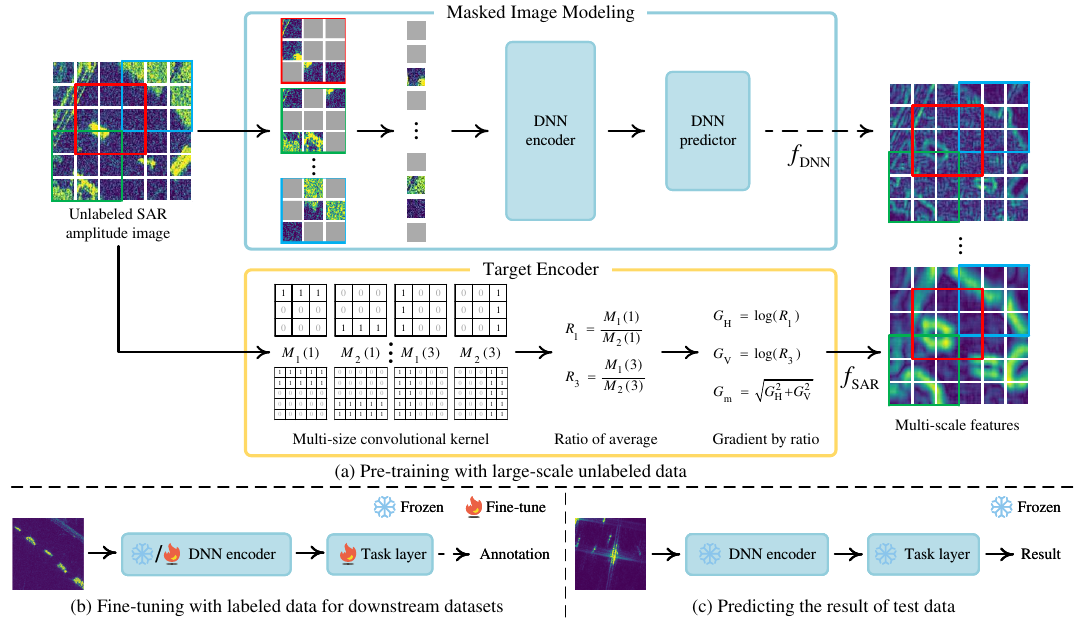}
\caption{Overall framework of \textcolor{black}{SAR-JEPA}. (a) \textcolor{black}{In the pretraining stage, joint-embedding predictive architecture for SAR automatic target recognition (SAR-JEPA)} uses local masked patches to predict the multi-scale SAR feature representations \textcolor{black}{$f_{\rm SAR}$} of unseen patches. The MIM structure uses the Vision Transformer (ViT) in MAE to extract deep features \textcolor{black}{$f_{\rm Deep}$} of masked patches. \textcolor{black}{Its} DNN predictor predicts the SAR features of unseen patches. The target encoder uses the GR method to map SAR images from pixel to feature space, thus extracting the target shapes and avoiding speckle noise interference in SAR images. Local patches and multi-scale features are designed for multi-scale small targets in remote sensing. \textcolor{black}{For downstream datasets, we fine-tune the DNN encoder with a task layer using labeled training data. DNN encoder weights are frozen or fine-tuned using various tuning settings. The weights of the other model modules are removed. (c) The trained model is used to predict the test result.}}
\label{fig3_framework}
\end{figure*}

The proposed \textcolor{black}{SAR-JEPA} is illustrated in Figure~\ref{fig3_framework}. Our \textcolor{black}{general} objective is as follows: given local masked patches, \textcolor{black}{predicting} feature representations in the SAR domain with respect to the model can interpret the SAR image by recognizing the contextual relationship between the target parts and the surrounding area.

As shown in Figure~\ref{fig3_framework}, the input is the single-channel SAR magnitude image because most target datasets are magnitude pictures. We randomly select different local patches and add mask tokens based on LoMaR~\citep{chen2022efficient}. ViTs~\citep{dosovitskiy2020image} in MAE are used as the DNN-encoder and DNN-predictor. Each local masked patch is fed into the MIM structure to predict the target feature of the mask tokens. \textcolor{black}{Importantly, the special SAR coherent imaging mechanism requires predicting high-quality target features instead of noisy pixels}. SAR coherent imaging results in multiplicative speckle noise in the magnitude image, and we use GR~\citep{dellinger2014sar,song2016sar} to suppress noise and extract the target shapes. \textcolor{black}{Moreover, we employ a multi-scale and local mask setting for various targets in the pretraining set.}

\subsection{Local Masked Patches}
We aim to learn the contextual relationship \textcolor{black}{between} various objects. However, such relationships usually remain in small areas because of the large scale and small target in remote sensing imagery. Therefore, we perform masking on the local patches following LoMaR. Although LoMaR is designed to improve computational efficiency, we \textcolor{black}{found that} this local setting is \textcolor{black}{suitable} for small targets in remote sensing. According to LoMaR, mask tokens and relative positional encoding~\citep{wu2021rethinking} are added to the encoder input and MAE structure to improve the feature representation. \textcolor{black}{These are important for LoMaR to avoid feature collapse.}

\subsection{Target Encoder}
\textcolor{black}{Owing} to the SAR imaging mechanism~\citep{ref1}, the image amplitude values contain multiplicative speckle noise. Speckle noise is related to the sensor resolution and is caused by the coherent superposition of different scatterers within a scattering cell. Therefore, reconstructing image pixels for SAR images \textcolor{black}{suffers from} noise interference and fails to learn a high-quality feature representation. This speckle noise problem can be solved using filtering and feature extraction. However, filtering has a conflict between noise reduction and image details, and we use feature extraction. Some studies have proposed \textcolor{black}{various} types of target feature in computer vision, \textcolor{black}{which are} mainly divided into traditional manual features~\citep{wei2022masked} and deep learning features~\citep{assran2023self}. However, these methods \textcolor{black}{cannot easily migrate directly} to SAR target recognition, and we combine well-established local description feature methods~\citep{dellinger2014sar,song2016sar,dong2020keypoint} in the SAR domain. 

\textbf{Gradient computation.}
The differential gradient is not a constant false alarm rate operator \textcolor{black}{because of the} multiplicative speckle noise in the SAR image. Previous studies~\citep{touzi1988statistical,bovik1988detecting} have shown that the computing ratio is more suitable for multiplicative noise. Here, we use the simplest ratio of average (ROA)~\citep{touzi1988statistical} to \textcolor{black}{avoid blurred} image details:
\begin{equation}\label{ROA}
R_i = \frac{M_1(i)}{M_2(i)},
\end{equation}
where $R_i$ denotes the average ratio at different directions and $M_1(i)$ and $M_2(i)$ denote the area averages on opposite sides of the current pixel along direction $i$. $i = 1$ is the horizontal direction, and $i = 3$ is the vertical direction. As shown in Figure~\ref{fig3_framework}, the area averages $M$ can be computed from the input image using four fixed convolution kernels\footnote{\textcolor{black}{We added 1e-2 to the input image to prevent zero values.}}. \textcolor{black}{Each convolution kernel extracts the sum of the region on one side of the pixel point, which is averaged to obtain $M$. By varying the sizes of the convolution kernel, multi-scale features $f_{\rm SAR}$ can be obtained.}

Then, the use of logarithms can solve the vertical gradient calculation~\citep{dellinger2014sar}, and the horizontal and vertical gradients are defined as follows:
\begin{equation}
\centering
\begin{split}
G_\text{H} &= log(R_1),\\
G_\text{V} &= log(R_3),
\end{split}
\end{equation}
where $G_\text{H}$ denotes the horizontal gradient and $G_\text{V}$ denotes the vertical gradient. The magnitude of the gradient is $G_\text{m} = \sqrt{G_\text{H}^2+G_\text{V}^2}$. We use gradient magnitude to construct multi-scale gradient features $f_{\rm SAR}$.

\textbf{Multi-Scale Feature.} 
SAR-HOG~\citep{song2016sar} discusses the kernel sizes for the MSTAR vehicle and provides the best single size for the MSTAR vehicle dataset. However, a single-sized kernel cannot accommodate various targets when the dataset is extended to different targets and sensors. Owing to the dynamic range of contextual information for various targets in remote sensing~\citep{li2023large}, a multi-scale feature is constructed with convolutional kernels of various sizes, \textcolor{black}{as depicted in} Figure~\ref{fig2_motivation}. 

\textcolor{black}{We set the kernel size $r$ equal to 5, 9, 13, and 17, and the whole convolution kernel is a square with an odd number $2r+1$. Using various kernel sizes, we have four gradient features, $G_{\rm m1}$, $G_{\rm m2}$, $G_{\rm m3}$, and $G_{\rm m4}$, and the multi-scale feature $f_{\rm SAR}={\rm cat}(G_{\rm m1}, G_{\rm m2}, G_{\rm m3}, G_{\rm m4})$ is a concatenate of four features in the feature channel.} As shown in Figure~\ref{fig2_motivation}, small-scale convolution kernels provide a finer extraction of small target contours, while large-scale convolution kernels focus on large targets, scene edges, and noise suppression.

\subsection{Implementation}
Given SAR images, we randomly sample several square windows at different spatial locations as local patches and mask each patch with mask tokens at a fixed percentage. Then, the \textcolor{black}{visible and unseen parts of} each local masked patch are fed into the MIM structure, whose encoder and predictor are applied with learnable relative positional encoding in the self-attention layer. Moreover, the entire SAR image is provided to the target encoder. The average values in four directions are obtained after multi-scale convolution in four directions. Then, the multi-scale SAR features are obtained after performing the ratio, logarithm, and magnitude operations. The loss function computes the mean squared error in the feature space between the DNN features $f_{\rm Deep}$ and SAR multi-scale features $f_{\rm SAR}$ for masked patches. In addition to the major improvement in the target encoder, there are two other minor \textcolor{black}{ones}.

\textbf{Data augmentation.} 
We follow the simple data augmentations in MAE and add a random contrast adjustment for SAR magnitude images.

\textbf{Decoder design.} 
Although some studies~\citep{chen2022efficient,wei2022masked} have shown that it is feasible to use a linear head as a decoder for SSL in RGB images, we find that a ViT with eight layers in the MAE is less disturbed by SAR image noise than a linear layer, thus allowing the encoder to extract deep semantic features of targets.

\section{Experiments}
\label{Experiments}
This section analyzes the proposed architecture. We \textcolor{black}{first describe the experimental setting for pretraining} and few-shot classification in Subsection \ref{Experimental Setting}. \textcolor{black}{Then, we discuss the effectiveness of the different modules in our architecture in Subsection \ref{Ablation Study}.} The proposed \textcolor{black}{SAR-JEPA} was compared with other methods in Subsections \ref{Results} and \ref{Visualization}. In the end, Subsection \ref{Scaling Experiment} discusses the scaling experiment, and SAR-JEPA shows improved performance as the amount of unlabeled data increases.

\subsection{Dataset and Experimental Settings}
\label{Experimental Setting}
\begin{table*}[!tb]
\footnotesize
\centering
\caption{Description of SAR datasets used for pretraining \textcolor{black}{and downstream tasks}. \# Target: Number of target categories. \# Scene: Number of scenes. Res.: Resolution. Large SAR imagery in the detection contains more target and scene types than the annotation.}
\label{table_dataset}
\renewcommand\arraystretch{1.1}
\begin{tabular}{cccccccl} 
\toprule
\multicolumn{1}{c}{Dataset} & Size & \# Target & \# Scene & Res. (m) & Band & Polarization & \multicolumn{1}{l}{Description} \\ 
\cmidrule(r){1-8}
MSAR & 28,499 & $\geq$4 & $\geq$6 & 1 & C & Quad & Ground and sea target detection dataset \\
SAR-Ship & 39,729 & $\geq$1 & $\geq$4 & 3$\sim$25 & C & Quad & Ship detection dataset in complex scenes \\
SARSim & 21,168 & 7 & 3 & 0.3 & X & Single & Vehicle simulation dataset \\
SAMPLE & 5,380 & 10 & 1 & 0.3 & X & Single & Vehicle simulation and measured~dataset \\ 
\cmidrule(lr){1-8}
\textcolor{black}{MSTAR} & 5,216 & 10 & 1 & 0.3 & X & Single & Fine-grained vehicle classification dataset \\
\textcolor{black}{FUSAR-Ship} & 9,830 & 10 & $\geq$5 & 1.1$\sim$1.7 & C & Double & Fine-grained ship classification dataset\\
\textcolor{black}{SAR-ACD} & 2,537 & 6 & 3 & 1 & C & Single & Fine-grained aircraft classification dataset \\
\bottomrule
\end{tabular}
\end{table*}

\begin{figure*}[!tb]
\centering
\includegraphics[width=17.5cm]{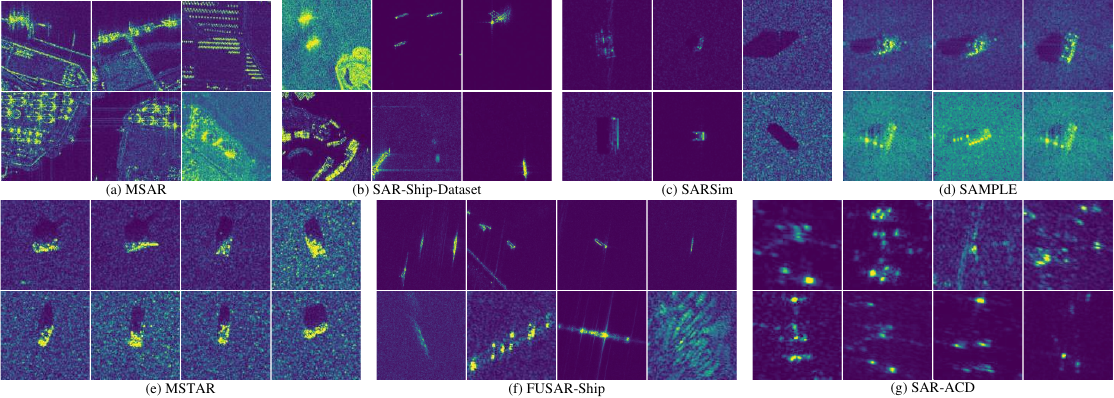}
\caption{\textcolor{black}{Datasets for the pretraining and downstream tasks.} Pretraining contains various targets, scenes, and sensors from MSAR, SAR-Ship-Dataset, SARSim, and SAMPLE. MSAR is the satellite-based dataset of ground and sea targets; SAR-Ship-Dataset is the satellite-based dataset of sea targets; SARSim is the multi-angle simulation dataset of vehicle targets; and SAMPLE is the simulated and measured dataset of vehicle targets. \textcolor{black}{We use MSTAR, FUSAR-Ship, and SAR-ACD datasets to evaluate the performance in recognizing different targets. MSTAR is a fine-grained vehicle dataset; FUSAR-Ship is a sea target dataset; and SAR-ACD is a fine-grained aircraft dataset.}}
\label{fig4_datasets}
\end{figure*}

We examine SSL performance using the following procedure. First, we perform SSL on the pretraining dataset without label information. Then, we fine-tune the pretrained model on few-shot SAR classification datasets. 

\textbf{Pretraining.}
Pretraining datasets are designed to provide rich information about targets, scenes, and sensors. \textcolor{black}{Therefore, we chose four datasets to build a pretraining dataset for SSL}. It covers popular targets (vehicles, ships, aircraft), complex backgrounds (ground and ocean), and various sensors (satellite-borne, airplane-borne, and simulation), as illustrated in Table~\ref{table_dataset} and Figure ~\ref{fig4_datasets}. 

MSAR~\citep{xia2022crtranssar,chen2022large} is a multi-class target detection dataset based on the Chinese HISEA-1 satellite in large-scale scenes. MSAR comprises 28,449 image slices with quad polarization. The scenes covered include airports, harbors, nearshore, islands, distant seas, and urban areas. Labeled target categories include aircraft, oil tanks, bridges, and ships. This dataset provides target samples in complex ground and sea scenes.

SAR-Ship~\citep{wang2019sar} dataset is a ship detection dataset in complex scenes based on Chinese Gaofen-3 and European Sentinel-1 satellites. The public version of this dataset contains 39,729 images from two satellites with various imaging modes and resolutions. The dataset provides ship targets of various sizes in complex ocean scenes, such as nearshore, distant seas, harbors, and islands.

SARSim~\citep{malmgren2017improving,kusk2016synthetic} is a fine-grained vehicle dataset created by Terma A/S, Denmark. The simulation system used for this dataset can generate X-band SAR images with resolutions ranging from 0.1 m to 0.3 m from \textcolor{black}{computer-aided design (CAD)} models. SARSim provides 21,168 vehicle samples in 7 categories (truck, car, motorbike, bus, tank, bulldozer, and pickup) and 3 scenes (grass, roads, and a mean of the two) with 7 imaging depression angles. This dataset provides fine-grained differences between targets. 

SAMPLE~\citep{lewis2019sar} is a fine-grained synthetic and measured paired vehicle dataset released by the Air Force Research Laboratory, USA. This dataset is simulated in X-band and 0.3 m resolution. The public version of this dataset provides 5,380 images of 10 categories of military vehicle targets at partial imaging angles, and 5 categories overlap with the public version of MSTAR. As shown in Figure~\ref{fig4_datasets}, it can be \textcolor{black}{observed} that the ground clutter in SAMPLE is more significant and realistic than in SARSim.

\textbf{Downstream tasks.}
We tested the pretrained model on three target recognition datasets to evaluate its ability for fine-grained semantic classification in the case of small samples with fine-tuning and linear probing. Based on MAE, fine-tuning is tuning all model parameters, and linear probing is tuning a linear layer with batch normalization while freezing the other parameters.

MSTAR~\citep{MSTAR} is the most commonly used SAR target classification dataset released by the Defense Advanced Research Projects Agency, USA, and it contains ten categories of vehicles with different angles, variants, and other imaging conditions. It has many variants of experimental settings, and we refer to~\citep{10283916} to adopt the most widely used ten-class classification settings to evaluate the fine-grained classification performance of SAR vehicles.

FUSAR-Ship~\citep{sun2022scan} contains 15 primary ship categories and many nonship targets based on the Gaofen-3 satellite in scenes, such as sea, land, coast, river, and island. Based on the experimental setting~\citep{wang2022sar}, we have ten types of ocean target slices from the original images, such as four fine-grained ships, bridges, and ocean scene slices. We use this dataset to evaluate the fine-grained classification performance of ocene targets.

SAR-ACD~\citep{sun2022scan} contains six types of aircraft based on the Gaofen-3 satellite in three civil airports: Shanghai Hongqiao Airport, Beijing Capital Airport, and Taiwan Taoyuan Airport. Because the released dataset does not separate the training and test data, we randomly select small samples as the training set and the others as the test set. We use this dataset to evaluate fine-grained aircraft target classification. The fine-grained recognition of aircraft targets is a more challenging task \textcolor{black}{owing} to the smooth surface of the aircraft, resulting in insignificant features. 

\textbf{Experimental settings.} 
Table~\ref{table: hyperparameter settings} \textcolor{black}{presents} the default setting. The pretraining is applied on 4 NVIDIA V100 GPUs with 200 epochs and 300 batch sizes, and the backbone model is ViT-Base (ViT-B). Other specific hyperparameters of each method use the default settings of their papers and codes. Compared to the training settings of MAE, we add ColorJitter\footnote{\textcolor{black}{ColorJitter is a Pytorch data augmentation method for adjusting an image's brightness, contrast, saturation, and hue. Considering the strong and weak scattering points in SAR magnitude images, adjusting contrast can effectively increase data diversity~\citep{geng2023target}. Therefore, we perform the contrast adjustment through a one-line Pytorch code ``transforms.ColorJitter (contrast=0.5)''.}} to increase the data richness, and all methods use the same data augmentation. \textcolor{black}{Because our goal is not to obtain better performance by tuning downstream hyperparameters,} all the models use the same training settings in downstream classification tasks. The few-shot learning setting is based on the Dassl toolbox~\citep{zhou2022domain} and is averaged over 10 random experiments. 

\begin{table}[!tb]\centering
\footnotesize
\centering
\caption{Experimental setting}
\label{table: hyperparameter settings}
\renewcommand\arraystretch{1.1}
\begin{tabular}{ll} 
\toprule
\multicolumn{2}{c}{\textbf{Pretraining setting}} \\
\cmidrule(){1-2}
Config & Value \\ 
\cmidrule(lr){1-2}
optimizer & AdamW \\
base learning rate & 1e-3 \\
weight decay & 0.05 \\
optimizer momentum & \textit{}$\beta_1,\beta_2 = 0.9,0.95$\\
batch size & 300 \\
epoch & 200 \\
learning rate schedule & cosine decay \\
warmup epoch & 20 \\
augmentation & ResizedCrop, HFlip, Contrast \\
\cmidrule(){1-2}
\multicolumn{2}{c}{\textbf{Classification settings}} \\
\cmidrule(){1-2}
Config & Value \\ 
\cmidrule(lr){1-2}
optimizer & AdamW~ \\
base learning rate & 1e-3 \\
weight decay & 5e-4 \\
optimizer momentum & \textit{}$\beta_1,\beta_2 = 0.9,0.999$~ \\
batch size & 50 \\
epoch & 40 \\
learning rate schedule & cosine decay~ \\
warmup epoch & 2 \\
warmup type & constant \\
warmup learning rate & 1e-5 \\
\bottomrule
\end{tabular}
\end{table}

\subsection{Ablation Study}
\label{Ablation Study}
\textcolor{black}{Here, we analyze three key points of the proposed method. The first compares the local (LoMaR) and global (MAE) mask methods and explains why we apply LoMaR with its modified ViT and the deep decoder of MAE as the backbone. Then, we discuss different target features (LPF, HOG, SAR-HOG, and ours) on the same backbone to show the need to suppress noise and target features. Finally, we discuss the role of MIM and PGCA in our architecture.}

\begin{table*}[!tb]
\centering
\footnotesize
\caption{\textcolor{black}{Discussion of the MIM framework for SAR-JEPA. MAE performs global masks, and LoMaR performs local masks. LoMaR-SAR is our modified LoMaR with decoder depth, and we use it as our SAR-JEPA backbone. The classification metric is the average accuracy (\%). \textbf{Bold} indicates the best result, and \underline{underline} indicates the next best result.}}
\label{table_backbone}
\renewcommand\arraystretch{1.1}
\begin{threeparttable}
\begin{tabular}{cccccccccc} 
\toprule
\multicolumn{10}{c}{\textbf{Fine-tuning}} \\
\cmidrule(){1-10}
\multicolumn{1}{c}{\multirow{2}{*}{Method}} & \multicolumn{3}{c}{MSTAR} & \multicolumn{3}{c}{FUSAR-Ship} & \multicolumn{3}{c}{SAR-ACD} \\
\cmidrule(lr){2-4} \cmidrule(lr){5-7} \cmidrule(lr){8-10}
 & 10-shot & 20-shot & 40-shot & 10-shot & 20-shot & 40-shot & 10-shot & 20-shot & 40-shot\\
\cmidrule(lr){1-10}
MAE~\citep{he2022masked} & \underline{50.6} & 61.3 & 69.5 & 71.7 & 75.4 & 78.2 & \underline{51.6} & \underline{57.0} & \underline{69.5} \\
LoMaR~\citep{chen2022efficient} & 45.6 & \underline{62.9} & \underline{77.0} & \underline{75.9} & \underline{80.2} & \underline{82.7} & 51.2 & 54.4 & 67.4\\
LoMaR-SAR\tnote{*} & \textbf{55.3} & \textbf{69.8} & \textbf{84.5} & \textbf{78.6} & \textbf{82.0} & \textbf{84.8} & \textbf{51.7} & \textbf{60.1} & \textbf{71.7}\\
\cmidrule(){1-10}
\multicolumn{10}{c}{\textbf{Linear probing}} \\
\cmidrule(){1-10}
\multicolumn{1}{c}{\multirow{2}{*}{Method}} & \multicolumn{3}{c}{MSTAR} & \multicolumn{3}{c}{FUSAR-Ship} & \multicolumn{3}{c}{SAR-ACD} \\
\cmidrule(lr){2-4} \cmidrule(lr){5-7} \cmidrule(lr){8-10}
 & 10-shot & 20-shot & 40-shot & 10-shot & 20-shot & 40-shot & 10-shot & 20-shot & 40-shot\\
\cmidrule(lr){1-10}
MAE~\citep{he2022masked} & \underline{53.7} & \underline{59.2} & \underline{63.7} & \underline{74.2} & \underline{78.1} & 80.3 & \underline{49.9} & \underline{56.9} & \underline{62.7} \\
LoMaR~\citep{chen2022efficient} & 39.3 & 48.9 & 55.6 & 70.5 & 77.9 & \underline{80.7} & 47.8 & 53.4 & 59.1\\
LoMaR-SAR\tnote{*} & \textbf{57.7} & \textbf{64.4} & \textbf{72.2} & \textbf{78.0} & \textbf{81.0} & \textbf{84.0} & \textbf{54.4} & \textbf{58.9} & \textbf{64.4}\\
\bottomrule
\end{tabular}
\begin{tablenotes}
\footnotesize
\item[*] {\scriptsize We replace the linear decoder of LoMaR with an eight-layer decoder in MAE to reduce the effect of noise in the reconstructed pixels by increasing the decoder depth. This improvement inspired us to thoroughly resolve SAR image noise with high-quality target features.}
\end{tablenotes}
\end{threeparttable}
\end{table*}

\begin{table*}[!tb]
\centering
\footnotesize
\caption{Results of various target features for \textcolor{black}{SAR-JEPA}. LPF: low pass filter. HOG: histogram of oriented gradient. GR: gradient-by-ratio. \textcolor{black}{The baseline is a pixel value that corresponds to LoMaR-SAR, as illustrated in Table~\ref{table_backbone}. LPF uses the default filter hyperparameters~\citep{liu2023pixmim}. SAR-HOG and two GR methods, whose kernel size can be adjusted, use the same multi-scale setting.} \textcolor{black}{$\rm{GR_{lin}}$: GR with linear kernel. $\rm{GR_{Gau}}$: GR with Gaussian kernel.} 
The classification metric is the average accuracy (\%). \textbf{Bold} indicates the best result, and \underline{underline} indicates the next best result.}
\label{table_target_feature}
\renewcommand\arraystretch{1.1}
\begin{threeparttable}
\begin{tabular}{cccccccccc} 
\toprule
\multicolumn{10}{c}{\textbf{Fine-tuning}} \\
\cmidrule(){1-10}
\multicolumn{1}{c}{\multirow{2}{*}{Target feature}} & \multicolumn{3}{c}{MSTAR} & \multicolumn{3}{c}{FUSAR-Ship} & \multicolumn{3}{c}{SAR-ACD} \\
\cmidrule(lr){2-4} \cmidrule(lr){5-7} \cmidrule(lr){8-10}
 & 10-shot & 20-shot & 40-shot & 10-shot & 20-shot & 40-shot & 10-shot & 20-shot & 40-shot\\
\cmidrule(lr){1-10}
Pixel & 55.3 & 69.8 & 84.5 & 78.6 & 82.0 & 84.8 & 51.7 & 60.1 & 71.7\\
LPF~\citep{liu2023pixmim} & 57.7 & 72.7 & 86.5 & 78.4 & 81.9 & 84.9 & 51.9 & 61.2 & 72.8\\
HOG~\citep{wei2022masked} & 30.2 & 32.4 & 40.3 & 58.1 & 63.3 & 66.6 & 51.7 & 57.9 & 64.7\\
SAR-HOG~\citep{song2016sar} & \underline{64.4} & 76.4 & \underline{90.5} & 76.7 & 80.8 & 85.3 & 51.4 & 60.0 & 73.4\\
\textbf{$\rm{GR_{Gau}}$}\tnote{*}~\citep{dellinger2014sar} & 62.8 & \underline{78.5} & 88.2 & \underline{80.6} & \underline{82.8} & \textbf{86.0} & \textbf{55.3} & \textbf{62.9} & \underline{73.0}\\
$\rm{GR_{lin}}$~\citep{song2016sar} & \textbf{70.3} & \textbf{82.1} & \textbf{91.6} & \textbf{81.3} & \textbf{83.3} & \underline{85.8} & \underline{54.8} & \underline{62.6} & \textbf{75.5} \\
\cmidrule(){1-10}
\multicolumn{10}{c}{\textbf{Linear probing}} \\
\cmidrule(){1-10}
\multicolumn{1}{c}{\multirow{2}{*}{Target feature}} & \multicolumn{3}{c}{MSTAR} & \multicolumn{3}{c}{FUSAR-Ship} & \multicolumn{3}{c}{SAR-ACD} \\
\cmidrule(lr){2-4} \cmidrule(lr){5-7} \cmidrule(lr){8-10}
 & 10-shot & 20-shot & 40-shot & 10-shot & 20-shot & 40-shot & 10-shot & 20-shot & 40-shot\\
\cmidrule(lr){1-10}
Pixel & 57.7 & 64.4 & 72.2 & 78.0 & 81.0 & 84.0 & 54.4 & 58.9 & 64.4\\
LPF~\citep{liu2023pixmim} & 56.1 & 62.6 & 68.1 & 78.7 & 82.0 & 84.6 & 54.9 & 59.5 & 65.2\\
HOG~\citep{wei2022masked} & 33.3 & 37.8 & 41.6 & 52.6 & 58.3 & 61.5 & 49.2 & 53.1 & 56.7\\
SAR-HOG~\citep{song2016sar} & 58.9 & 68.1 & 77.1 & 79.3 & 81.8 & 85.0 & 54.6 & 59.3 & 66.5\\
$\rm{GR_{Gau}}$\tnote{*}~\citep{dellinger2014sar} & \underline{62.8} & \underline{72.3} & \underline{77.6} & \textbf{82.0} & \textbf{82.9} & \textbf{85.7} & \underline{54.7} & \textbf{60.9} & \underline{66.6}\\
$\rm{GR_{lin}}$~\citep{song2016sar} & \textbf{67.7} & \textbf{75.1} & \textbf{81.6} & \underline{80.6} & \underline{82.7} & \underline{85.2} & \textbf{56.8} & \underline{60.4} & \textbf{66.8}\\
\bottomrule
\end{tabular}
\begin{tablenotes}
\footnotesize
\item[*] {\textcolor{black}{\scriptsize We finally chose the linear kernel $\rm{GR_{lin}}$ for SAR-JEPA because it requires considerable time to fine-tune the Gaussian kernel $\rm{GR_{Gau}}$ parameters in pretraining. we set the Gaussian standard deviation to $\sigma=0.3*((2r+1-1)*0.5-1)+0.8$. This implies that the standard deviation is related to the kernel size instead of a fixed parameter.}}
\end{tablenotes}
\end{threeparttable}
\end{table*}

\begin{figure*}[!tb]
\centering
\footnotesize
\includegraphics[width=16.0cm]{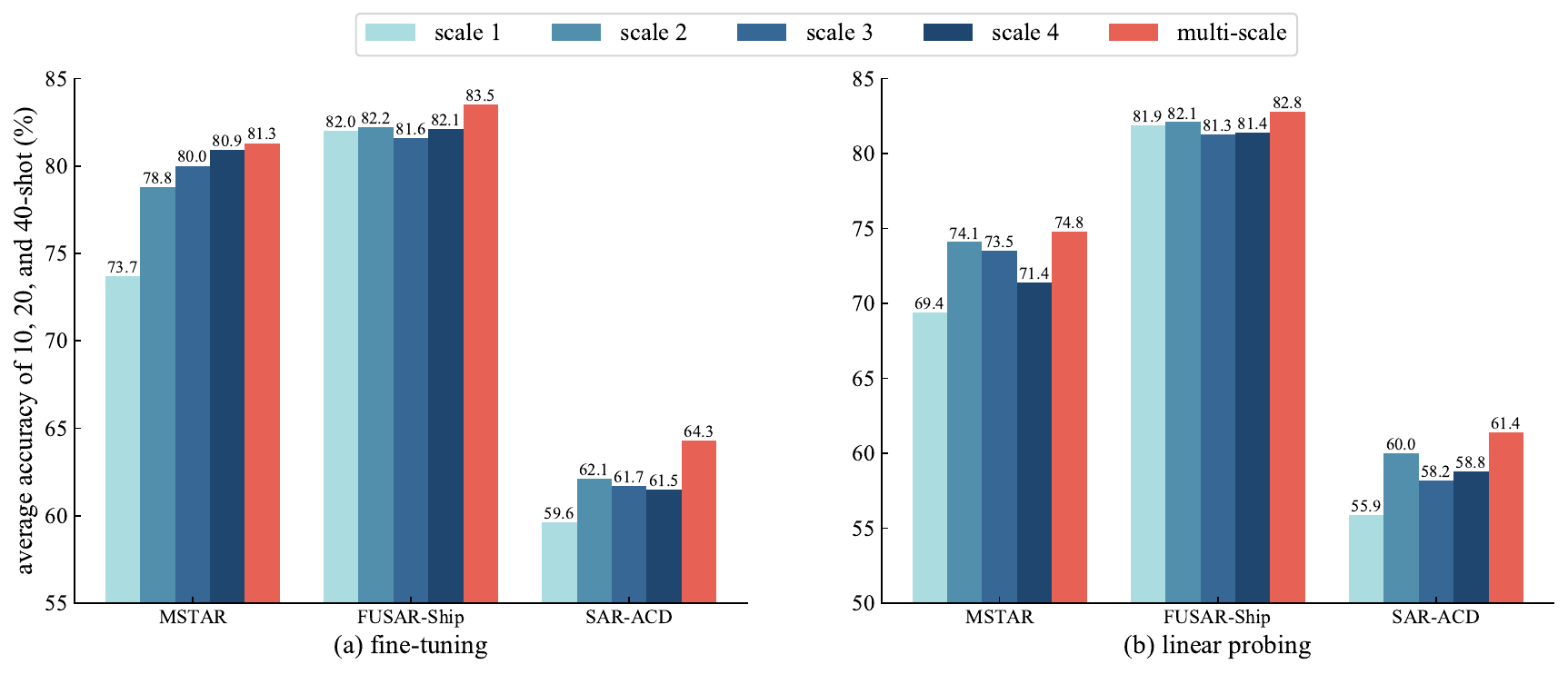}
\caption{Multi-scale kernel settings for $\rm{GR_{lin}}$. Here, the scale 1/2/3/4 has $r$ equal to 5/9/13/17, and the multi-scale \textcolor{black}{concat all four scales in the feature channel}. Multi-scale is more suitable than single scale because of its various targets in remote sensing images.}
\label{fig_analysis_multiscale}
\end{figure*}

\begin{table*}[!tb]
\centering
\footnotesize
\caption{\textcolor{black}{Ablation study of our SAR-JEPA combines two architectures (MIM and PGCA). The classification metric is the average accuracy (\%). \textbf{Bold} indicates the best result, and \underline{underline} indicates the next best result.}}
\label{table_sarjepa}
\renewcommand\arraystretch{1.1}
\begin{threeparttable}
\begin{tabular}{cccccccccc} 
\toprule
\multicolumn{10}{c}{\textbf{Fine-tuning}} \\
\cmidrule(){1-10}
\multicolumn{1}{c}{\multirow{2}{*}{Method}} & \multicolumn{3}{c}{MSTAR} & \multicolumn{3}{c}{FUSAR-Ship} & \multicolumn{3}{c}{SAR-ACD} \\
\cmidrule(lr){2-4} \cmidrule(lr){5-7} \cmidrule(lr){8-10}
 & 10-shot & 20-shot & 40-shot & 10-shot & 20-shot & 40-shot & 10-shot & 20-shot & 40-shot\\
\cmidrule(lr){1-10}
MIM\tnote{*} & 55.3 & 69.8 & 84.5 & \underline{78.6} & \underline{82.0} & \underline{84.8} & 51.7 & \underline{60.1} & 71.7\\
PGCA\tnote{**} & \underline{56.7} & \underline{71.4} & \underline{88.1} & 76.4 & 80.4 & 83.8 & \underline{51.8} & 60.0 & \underline{73.7}\\
SAR-JEPA & \textbf{70.3} & \textbf{82.1} & \textbf{91.6} & \textbf{81.3} & \textbf{83.3} & \textbf{85.8} & \textbf{54.8} & \textbf{62.6} & \textbf{75.5}\\
\cmidrule(){1-10}
\multicolumn{10}{c}{\textbf{Linear probing}} \\
\cmidrule(){1-10}
\multicolumn{1}{c}{\multirow{2}{*}{Method}} & \multicolumn{3}{c}{MSTAR} & \multicolumn{3}{c}{FUSAR-Ship} & \multicolumn{3}{c}{SAR-ACD} \\
\cmidrule(lr){2-4} \cmidrule(lr){5-7} \cmidrule(lr){8-10}
 & 10-shot & 20-shot & 40-shot & 10-shot & 20-shot & 40-shot & 10-shot & 20-shot & 40-shot\\
\cmidrule(lr){1-10}
MIM\tnote{*} & 57.7 & 64.4 & 72.2 & \underline{78.0} & 81.0 & 84.0 & \underline{54.4} & \underline{58.9} & 64.4\\
PGCA\tnote{**} & \underline{61.0} & \underline{68.0} & \underline{74.7} & 77.4 & \underline{81.6} & \underline{84.2} & 52.7 & 58.3 & \underline{66.1}\\
SAR-JEPA & \textbf{67.7} & \textbf{75.1} & \textbf{81.6} & \textbf{80.6} & \textbf{82.7} & \textbf{85.2} & \textbf{56.8} & \textbf{60.4} & \textbf{66.8}\\
\bottomrule
\end{tabular}
\begin{tablenotes}
\footnotesize
\item[*] {\scriptsize MIM is referred to as LoMaR-SAR in Table~\ref{table_backbone}, which is also the MIM module of our SAR-JEPA.}
\item[**] {\scriptsize Due to PGCA~\citep{datcu2023explainable} using the features of complex SAR image as target signals, we follow its ideas and use our target encoder. Here, PGCA is the contrastive learning between deep features and our multi-scale gradient features $\rm{GR_{lin}}$ to test the effectiveness of the target feature as a guided signal. It removes the MIM task from SAR-JEPA.}
\end{tablenotes}
\end{threeparttable}
\end{table*}

\textbf{LoMaR \emph{vs.} MAE.} 
\textcolor{black}{Because most targets are small objects in remote sensing images, we prefer masking in local areas to learn the contextual relationships near the target better. Compared with MAE, which masks the entire image, LoMaR randomly selects multiple local squares for masking. In addition, LoMaR uses relative positional encoding, input with mask tokens, and linear decoder. According to LoMaR, the first two are indispensable to prevent feature collapse. When we remove any of them, pretraining in SAR images with LoMaR and LoMaR-SAR has performance degradation and feature collapse.}

\textcolor{black}{LoMaR uses a linear decoder to accelerate the computation. However, the linear layer is unsuitable for SAR images because the SAR image noise easily affects the encoder features through the loss function. In our early experiments, LoMaR performed better than MAE on small batches of high-quality pretraining data. When we extended the pretraining set to various quality datasets, the performance of LoMaR became unstable. This phenomenon inspired us to improve the decoder and target features. As shown in Table~\ref{table_backbone}, LoMaR-SAR, which replaces the linear decoder of LoMaR with an eight-layer decoder in MAE\footnote{\textcolor{black}{However, deeper decoders can be detrimental to the encoder performance. For example, the 12-layer decoder performs worse than the 8-layer in the MAE paper.}}, exhibits the best result. Therefore, we use LoMaR-SAR as the MIM module for SAR-JEPA. The target encoder for SAR-JEPA is discussed below. It requires unique designs for SAR images and is important to our architecture.}

\textbf{Target feature selection:} 
In Table~\ref{table_target_feature}, we compare the results of the pixel value, low pass filter (LPF)~\citep{liu2023pixmim}, SAR-HOG~\citep{song2016sar}, and GR~\citep{dellinger2014sar}. All the changes are made only in the target features, whereas the other settings remain unchanged. \textcolor{black}{The Pixel shown in Table~\ref{table_target_feature} is LoMaR-SAR used as baseline}. The LPF can mitigate noise interference, but we observe a slight drop of LPF compared to Pixel on the linear probing result of MSTAR. Therefore, inspired by the HOG feature in MaskFeat, we prefer to obtain high-quality target features by suppressing noise.

Although HOG features \textcolor{black}{are effective} in FG-MAE for SAR scene-level SSL, we \textcolor{black}{observe that} HOG features are unsuitable for target-level SSL, as shown in Table~\ref{table_target_feature}. This result is because speckle noise can cause the gradient computation in small target regions to have many noise points, \textcolor{black}{thereby significantly} affecting the quality of the target features. Inspired by the physics-guided contrastive architecture using SAR physics features, we investigate the HOG feature variant in the SAR domain, that is, SAR-HOG. The core of SAR-HOG is the change in the gradient calculation from the differential gradient to the ratio gradient. As shown in Table~\ref{table_target_feature}, we observe a significant improvement compared to HOG owing to this modification. However, the sampling method of HOG may lead to the loss of some small target scattering points, such as the discrete points of an aircraft or a small boat. Therefore, we focus on the gradient information, that is, the target edges. Moreover, basic information, such as the shape and edges of an object, is a generic property~\citep{9193980,9523773,shi2020informative}.

We consider two \textcolor{black}{types} of GR in SAR. $\rm{GR_{lin}}$ is computes gradient using a linear kernel~\citep{song2016sar} whose $M$ value is equal to 1. $\rm{GR_{Gau}}$ computes gradient using a Gaussian kernel~\citep{dellinger2014sar} whose $M$ value obeys a Gaussian distribution. \textcolor{black}{Table~\ref{table_target_feature} shows that the two calculation methods have their advantages for various target categories and have similar performance in most cases. The Gaussian kernel can act as a filter, but the parameters must be adjusted to avoid excessive blurring of the target edges. Because a pretrained model requires tens of hours to several days, we prefer to use simple and effective methods with scaling data and parameters.} Therefore, we use a linear kernel without spending too much time tuning the parameters of the Gaussian kernel at \textcolor{black}{various} scales. 

Another important thing is that we propose a multi-scale setting instead of single-scale features for various targets included in the pretraining set. Multi-scale can improve the feature representation of various targets in remote sensing. Figure~\ref{fig_analysis_multiscale} \textcolor{black}{illustrates} the results of the various kernel settings. Scale settings have various effects: smaller scales can better extract small target edges, while larger scales are more obvious to noise suppression. Hence, learning target features at various scales is more advantageous than a single scale.

\textcolor{black}{\textbf{SAR-JEPA} is a combination of MIM and PGCA ideas, as shown in Figure~\ref{fig2_motivation}. In particular, we designed LoMaR-SAR and multi-scale gradient features as a concrete implementation of the MIM and target encoder module. Table~\ref{table_sarjepa} demonstrates the equal importance and mutual reinforcement of the two modules. They focus on the pretext task and signal quality of SSL driven by the data. MIM can provide PGCA using a challenging pretext task of learning features with contextual information in the image. PGCA can improve the quality of the features by introducing domain knowledge into the embedding of the target encoder. Combining MIM and PGCA, SAR-JEPA learns high-quality contextual features for low-quality noisy SAR data.}

\subsection{Comparisons with other methods}
\label{Results}
\begin{table*}[!tb]
\centering
\footnotesize
\caption{Fine-grained classification results of various methods. The classification metric is the average accuracy (\%). \textbf{Bold} indicates the best result, and \underline{underline} indicates the next best result.}
\label{table_result}
\renewcommand\arraystretch{1.1}
\begin{tabular}{ccccccccc} 
\toprule
\multicolumn{9}{c}{\textbf{MSTAR}} \\
\cmidrule(){1-9}
\multicolumn{1}{c}{\multirow{2}{*}{Method}} & \multicolumn{1}{c}{\multirow{2}{*}{Mask area}} & \multicolumn{1}{c}{\multirow{2}{*}{Target feature}} & \multicolumn{3}{c}{Fine-tuning} & \multicolumn{3}{c}{Linear probing} \\ 
\cmidrule(lr){4-6} \cmidrule(lr){7-9}
 &  &  & 10-shot & 20-shot & 40-shot & 10-shot & 20-shot & 40-shot \\ 
\cmidrule(lr){1-9}
ImageNet & - & - & 16.7 & 32.0 & 40.7 & 45.0 & 56.3 & \underline{66.8} \\
MAE~\citep{he2022masked} & Global & Pixel value & \underline{50.6} & 61.3 & 69.5 & \underline{53.7} & \underline{59.2} & 63.7 \\
FG-MAE~\citep{wang2023feature} & Global & HOG & 50.0 & 58.7 & 70.0 & 43.5 & 50.7 & 56.4 \\
I-JEPA~\citep{assran2023self} & Global & Deep feature & 24.5 & 33.4 & 44.6 & 30.9 & 38.6 & 44.7 \\
LoMaR~\citep{chen2022efficient} & Local & Pixel value & 45.6 & \underline{62.9} & \underline{77.0} & 39.3 & 48.9 & 55.6 \\ 
\textbf{\textcolor{black}{SAR-JEPA}} & Local & \textbf{$\rm{GR_{lin}}$} & \textbf{70.3} & \textbf{82.1} & \textbf{91.6} & \textbf{67.7} & \textbf{75.1} & \textbf{81.6} \\
\cmidrule(){1-9}
\multicolumn{9}{c}{\textbf{FUSAR-ship}} \\
\cmidrule(){1-9}
\multicolumn{1}{c}{\multirow{2}{*}{Method}} & \multicolumn{1}{c}{\multirow{2}{*}{Mask area}} & \multicolumn{1}{c}{\multirow{2}{*}{Target feature}} & \multicolumn{3}{c}{Fine-tuning} & \multicolumn{3}{c}{Linear probing} \\ 
\cmidrule(lr){4-6} \cmidrule(lr){7-9}
 &  &  & 10-shot & 20-shot & 40-shot & 10-shot & 20-shot & 40-shot \\ 
\cmidrule(lr){1-9}
ImageNet & - & - & 54.0 & 57.0 & 59.7 & \underline{74.6} & \underline{79.6} & \underline{83.6} \\
MAE~\citep{he2022masked} & Global & Pixel value & 71.7 & 75.4 & 78.2 & 74.2 & 78.1 & 80.3 \\
FG-MAE~\citep{wang2023feature} & Global & HOG & 65.5 & 73.7 & 76.6 & 69.7 & 74.3 & 77.7 \\
I-JEPA~\citep{assran2023self} & Global & Deep feature & 38.5 & 38.9 & 49.4 & 36.5 & 42.2 & 46.8 \\
LoMaR~\citep{chen2022efficient} & Local & Pixel value & \underline{75.9} & \underline{80.2} & \underline{82.7} & 70.5 & 77.9 & 80.7 \\ 
\textbf{\textcolor{black}{SAR-JEPA}} & Local & \textbf{$\rm{GR_{lin}}$} & \textbf{81.3} & \textbf{83.3} & \textbf{85.8} & \textbf{80.6} & \textbf{82.7} & \textbf{85.2} \\
\cmidrule(){1-9}
\multicolumn{9}{c}{\textbf{SAR-ACD}} \\
\cmidrule(){1-9}
\multicolumn{1}{c}{\multirow{2}{*}{Method}} & \multicolumn{1}{c}{\multirow{2}{*}{Mask area}} & \multicolumn{1}{c}{\multirow{2}{*}{Target feature}} & \multicolumn{3}{c}{Fine-tuning} & \multicolumn{3}{c}{Linear probing} \\ 
\cmidrule(lr){4-6} \cmidrule(lr){7-9}
 &  &  & 10-shot & 20-shot & 40-shot & 10-shot & 20-shot & 40-shot \\ 
\cmidrule(lr){1-9}
ImageNet & - & - & 22.6 & 26.8 & 42.7 & 34.8 & 44.2 & 54.2 \\
MAE~\citep{he2022masked} & Global & Pixel value & \underline{51.6} & \underline{57.0} & \underline{69.5} & \underline{49.9} & \underline{56.9} & \underline{62.7} \\
FG-MAE~\citep{wang2023feature} & Global & HOG & 50.4 & 54.2 & 62.9 & 47.5 & 51.5 & 57.0 \\
I-JEPA~\citep{assran2023self} & Global & Deep feature & 45.0 & 51.7 & 63.3 & 42.7 & 51.0 & 60.0 \\
LoMaR~\citep{chen2022efficient} & Local & Pixel value & 51.2 & 54.4 & 67.4 & 47.8 & 53.4 & 59.1 \\ 
\textbf{\textcolor{black}{SAR-JEPA}} & Local & \textbf{$\rm{GR_{lin}}$} & \textbf{54.8} & \textbf{62.6} & \textbf{75.5} & \textbf{56.8} & \textbf{60.4} & \textbf{66.8} \\
\bottomrule
\end{tabular}
\end{table*}

\begin{figure}[!tb]
\centering
\footnotesize
\includegraphics[width=8.1cm]{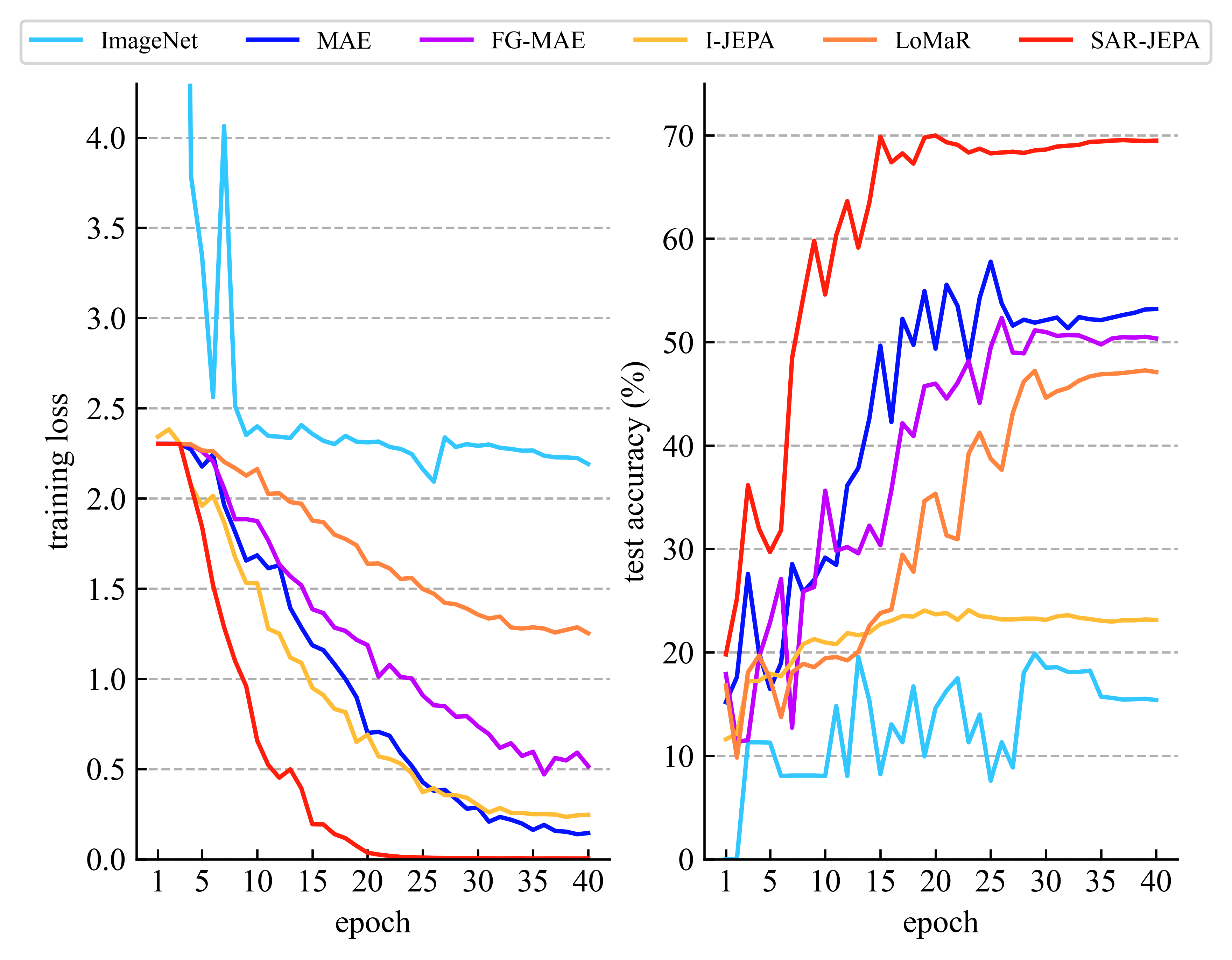}
\caption{\textcolor{black}{Training loss and test accuracy curves for one fine-tuning of MSTAR 10-shot (Table~\ref{table_result}). From the figure, effective SSL decreases the training loss rapidly and converges to a more generalized region.}}
\label{fig_training_curve}
\end{figure}

As shown in Table~\ref{table_result}, we evaluated SSL methods in few-shot classification tasks. The comparison methods employed are ImageNet, MAE~\citep{he2022masked}, Feature Guided MAE (FG-MAE)~\citep{wang2023feature}, I-JEPA~\citep{assran2023self}, LoMaR~\citep{chen2022efficient}, and our \textcolor{black}{SAR-JEPA}. Table~\ref{table_result} illustrates the difference in the mask area and target features. The ImageNet migrated the supervised pretraining weights on the ImageNet for initialization. 

Table~\ref{table_result} \textcolor{black}{presents the results of various pretrained models on the fine-grained classification of the SAR target. Owing} to the large difference between SAR and RGB images, ImageNet weights cannot easily obtain effective features with small samples when fine-tuning, and using linear probing can demonstrate a better result with an unsuitable initialization weight. MAE trained on the SAR dataset performs better than ImageNet with fine-tuning, but the speckle noise affects its representation quality. HOG is not an effective solution to speckle noise because multiplicative noise leads to virtual dots in the strong-scattering region. This results in a performance degradation in FG-MAE with HOG features, \textcolor{black}{and FG-MAE converges slower than MAE, as shown in Figure~\ref{fig_training_curve}}. Our method is a special form of the I-JEPA~\citep{assran2023self} method in SAR. However, I-JEPA has the feature collapse, in which the learnable predictor fails to learn a good feature space under speckle noise. \textcolor{black}{As depicted in Figure~\ref{fig_training_curve}, I-JEPA converges to a region with poor test performance.} \textcolor{black}{Figure~\ref{fig_training_curve} shows that effective SSL rapidly decreases training loss and converges to a more generalized region}

Another interesting phenomenon is the range of contexts, where it can be found that LoMaR with local reconstruction can achieve better fine-tuning results on vehicles and ships than MAE. This result illustrates the need to consider the range of contextual information in remote sensing images. However, because of a linear decoder, LoMaR is more susceptible to the effects of speckle noise in pixels, leading to \textcolor{black}{unstable performance illustrated in Table~\ref{table_result} and a slow convergence rate in Figure~\ref{fig_training_curve}}. 

Although the proposed method achieves better results on the aircraft dataset, the results are not as good as those of vehicles and ships. \textcolor{black}{In particular, fine-tuning at SAR-ACD 10-shot is worse than the linear probing, indicating slight overfitting phenomena.} This is because the aircraft reflects electromagnetic waves with its streamlined structure, resulting in SAR images that may lack the complete aircraft structure, with only some strong-scattering points and shadow edges. \textcolor{black}{Moreover, our pretraining did not include many aircraft target samples.} Consequently, the generic representation in SAR images still needs to be further explored.

\subsection{Visualization}
\label{Visualization}
\begin{figure*}[!htb]
\centering
\footnotesize
\includegraphics[width=16.0cm]{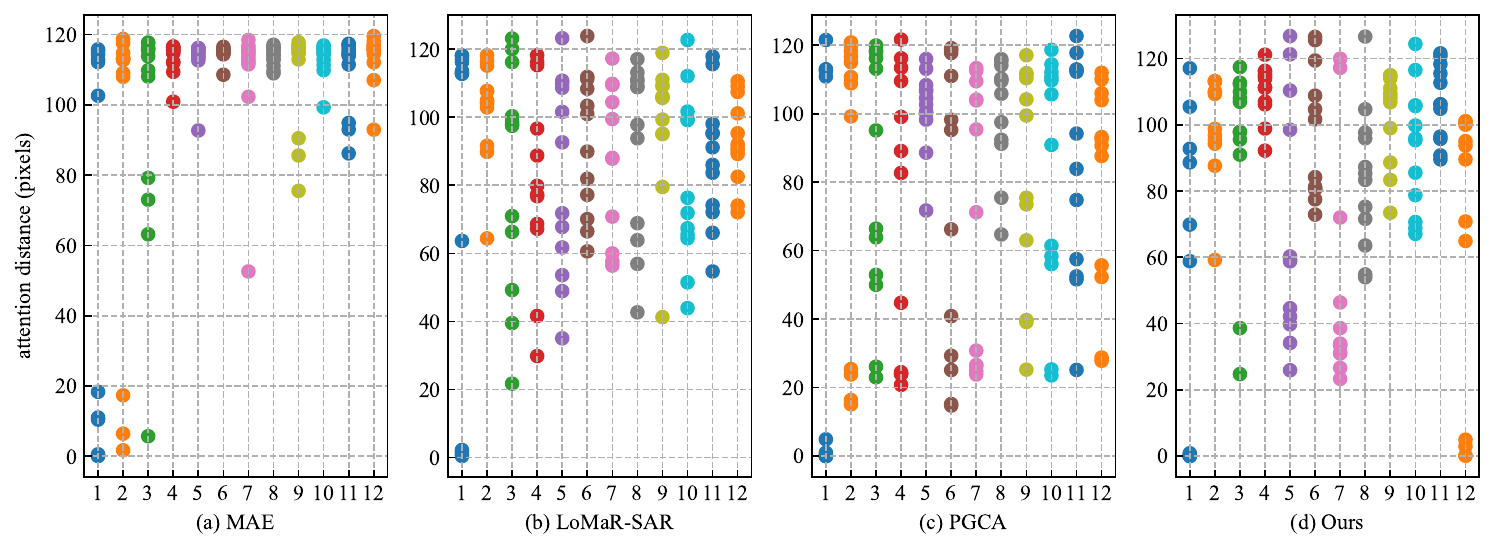}
\caption{The averaged attention distance of ViT using various methods. The X-axis is the attention head, that is, the layer number. \textcolor{black}{Attention distance~\citep{dosovitskiy2020image, xie2023revealing} is computed based on example pictures (we use 16 example pictures) by averaging the distance between the query pixel and all other pixels, weighted by the attention weight~\citep{dosovitskiy2020image}. A larger attention distance indicates the ability to focus on the correlation of two more distant pixels in an image.} MAE focuses on global information, and other methods focus on local and global attention ranges. This Figure shows that using pixel-level features as a pretext task with a local image scale can enhance the diversity of Vit modeling with SAR images.}
\label{visual_attention_distance}
\end{figure*}
Compared to supervised pretraining and contrastive learning, which have a local attention range in the high layer, ViT with MIM has various attention ranges in different layers~\citep {xie2023revealing}. Figure~\ref{visual_attention_distance} shows that this characteristic of MIM is related to the data property and the pixel-level pretext task. 
We first observe that the deep layers of MAE mainly focus on global information owing to large SAR image scenes, which differs from the modeling properties of MIM with ImageNet. Then, LoMaR-SAR with local reconstruction can model the local information. This phenomenon illustrates that local information is easily lost in remote sensing images, and local reconstruction can better capture contextual details. In addition, PGCA, which removes the mask in \textcolor{black}{SAR-JEPA} and performs a pixel-level contrast learning task, can obtain local information. This suggests that the property of modeling local and global information originates from the pixel-level pretext task. Finally, our \textcolor{black}{SAR-JEPA} with integrated MIM and PGCA can model local and global features. It can be observed that the addition of MIM makes SAR-JEPA concentrate more on high-level global contextual information and more diminished attention to local edge features compared to PGCA. \textcolor{black}{Therefore, the differences in the properties between SAR and RGB images make the same SSL method show various properties, which is a great challenge for the SSL methods in SAR.}

\subsection{Scaling Experiment}
\label{Scaling Experiment}
\begin{figure*}[!tb]
\centering
\footnotesize
\includegraphics[width=16.0cm]{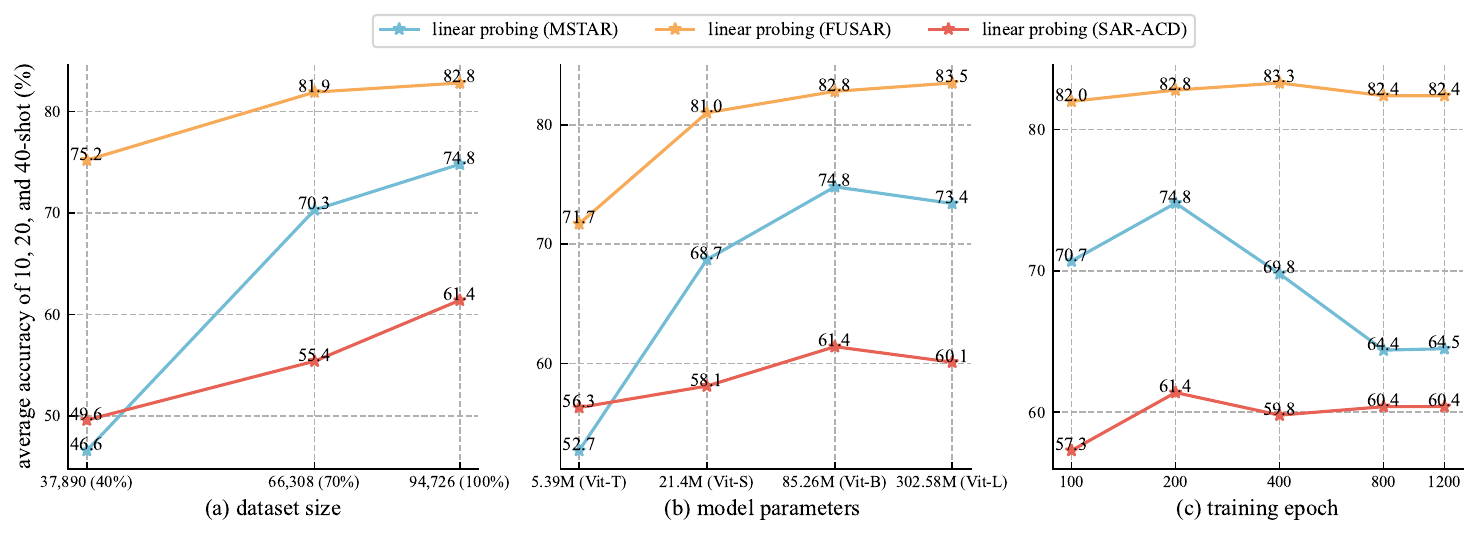}
\caption{Scaling experiment of \textcolor{black}{SAR-JEPA} in dataset size, model parameters, and training epoch. We observe that \textcolor{black}{SAR-JEPA} can benefit from these three \textcolor{black}{viewpoints. However, owing} to the limitation of dataset size, a large model and long training epochs lead to overfitting and reduced linear probing performance.}
\label{fig_scalability}
\end{figure*}

MIM can benefit from scaling data, model parameters, and computational resources, but it suffers from overfitting when using small data and long training epochs in large models~\citep{xie2023data}. Therefore, we evaluate the scalability of \textcolor{black}{SAR-JEPA} with linear probing performance on SAR data. Figure~\ref{fig_scalability} illustrates the scaling experiment for the dataset size, model parameters, and training epoch.

\textbf{Dataset size.} 
A significant performance increase is observed in small-sample classification as data volume increases. This phenomenon suggests that by extracting high-quality features as guide signals, SSL performance can be improved by increasing the number of noisy SAR images. Moreover, this result indicates that SSL does not reach saturation in the dataset size and has a significant potential for the foundation model to be exploited.

\textbf{Model parameters.} 
As the model parameters increase, we observe an increase in representation quality. However, Vit-Large suffered from overfitting problems because of insufficient dataset size and exhibited reduced performance in the fine-grained classification of vehicles and aircraft. 

\textbf{Training epoch.} 
Similarly, too-long training epochs result in overfitting of Vit-Base; consequently, our default training setting is 200 epochs. Therefore, the next step is to collect more SAR target data to exploit the potential of SSL and foundation models.

\section{Conclusion}
\label{Conclusion}
\textcolor{black}{In this study, we investigate SSL for SAR ATR using our SAR-JEPA. SAR-JEPA combines two architectures to perform mask image modeling in a gradient feature space for SAR images. By leveraging local patch masking and SAR domain features}, we extend the applicability of MIM to SAR imagery, which is a type of noisy remote sensing data. Through extensive experimentation and analysis, we demonstrate the effectiveness of the proposed \textcolor{black}{SAR-JEPA} in obtaining high-quality representations for SAR ATR across various target categories and image conditions. This study highlights the importance of using SSL to exploit the increasing availability of SAR data. In the future, we will further extend the data and computational resources and provide SSL with the potential to achieve a foundation model for various downstream SAR ATR tasks across targets, scenes, and sensors. 

\textcolor{black}{However, it is noteworthy that our pretraining dataset needs to scale extensively, like ImageNet, to exploit the potential of SSL for SAR ATR. To the best of our knowledge, other SAR datasets can be further added to the pretraining, such as vehicle datasets Gotcha~\citep{dungan2012wide}, SIVED~\citep{lin2023sived}, ship datasets AIR-SARSHIP~\citep{xian2019air}, HRSID~\citep{wei2020hrsid}, aircraft dataset SADD~\citep{zhang2022sefepnet}, and building datasets FUSAR-Map~\citep{9369836}, OGSOD~\citep{wang2023category}. However, scaling the SAR target data volume close to the million scale of ImageNet is a challenging problem. Another direction is to use the simulated data for SSL. For example, formula-driven supervised learning~\citep{kataoka2022pre} does not use any natural images for SSL in computer vision and learns different fractal structures.} In addition, more studies are required for other foundation model architectures, such as ConvNeXt. Although this study focuses primarily on feature representation across various targets, it is essential to explore and discuss various tasks, including detection and segmentation. In the future, we plan to enhance the proposed scheme by improving the data collection methods, refining the model architectures, and exploring novel pretraining target features. Our ultimate goal is to successfully apply the foundation model with SSL to SAR target recognition.










\bibliographystyle{cas-model2-names}

\bibliography{ref}



\end{document}